\pdfoutput=1

\documentclass[11pt]{article}

\usepackage{EMNLP2022}

\usepackage{times}

\usepackage{latexsym}
\usepackage{microtype}
\usepackage{amsmath, amssymb}
\usepackage{booktabs}
\usepackage{graphicx}
\usepackage{multirow}
\usepackage[noend]{algpseudocode}
\usepackage{amsmath}
\usepackage{algorithmicx,algorithm}
\usepackage[T1]{fontenc}

\usepackage[utf8]{inputenc}

\usepackage{microtype}

\usepackage{inconsolata}

\title{On the Off-Target Problem of Zero-Shot Multilingual \\ Neural Machine Translation}

\let\oldtheenumi=\thefootnote

\author{ Liang Chen$^1$ \quad Shuming Ma$^2$\quad Dongdong Zhang$^2$ \quad Furu Wei$^2$ \quad Baobao Chang$^1$\footnotemark[2]  \\ National Key Laboratory for Multimedia Information Processing, Peking University$^1$  \\ Microsoft Research$^2$ \\ \texttt{leo.liang.chen@outlook.com} \quad
 \texttt{chbb@pku.edu.cn} \\ \texttt{\{shumma,dozhang,fuwei\}@microsoft.com}}

\renewcommand{\thefootnote}{\fnsymbol{footnote}}

\begin{document}
\maketitle
\begin{abstract}
While multilingual neural machine translation has achieved great success, it suffers from the off-target issue, where the translation is in the wrong language. This problem is more pronounced on zero-shot translation tasks. In this work, we find that failing in encoding discriminative target language signal will lead to off-target and a closer lexical distance (i.e., KL-divergence) between two languages' vocabularies is related with a higher off-target rate. We also find that solely isolating the vocab of different languages in the decoder can alleviate the problem. Motivated by the findings, we propose Language Aware Vocabulary Sharing (LAVS), a simple and effective algorithm to construct the multilingual vocabulary, that greatly alleviates the off-target problem of the translation model by increasing the KL-divergence between languages. We conduct experiments on a multilingual machine translation benchmark in 11 languages. Experiments show that the off-target rate for 90 translation tasks is reduced from 29\% to 8\%, while the overall BLEU score is improved by an average of 1.9 points without extra training cost or sacrificing the supervised directions' performance. We release the code at \href{https://github.com/PKUnlp-icler/Off-Target-MNMT}{https://github.com/PKUnlp-icler/Off-Target-MNMT} for reproduction.

\footnotetext[2]{Corresponding author.}

\end{abstract}

\renewcommand{\thefootnote}{\oldtheenumi}

\section{Introduction}

Multilingual NMT makes it possible to do the translation among multiple languages using only one model, even for zero-shot directions~\citep{johnson-etal-2017-googles,aharoni-etal-2019-massively}. It has been gaining increasing attention since it can greatly reduce the MT system's deployment cost and enable knowledge transfer among different translation tasks, which is especially beneficial for low-resource languages. 
Despite its success, off-target is a harsh and widespread problem during zero-shot translation in existing multilingual models. For the zero-shot translation directions, the model translates the source sentence to a wrong language, which severely degrades the system's credibility. As shown in Table~\ref{tab:otr}, the average off-target rate on 90 directions is 29\% and even up to 95\% for some language pair (tr->gu) on WMT'10 dataset.

Researchers have been noticing and working on solving the problem from different perspectives. For model trained on English-centric dataset, a straight forward method is to add pseudo training data on the zero-shot directions through back-translation~\citep{gu-etal-2019-improved,zhang-etal-2020-improving}. Adding pseudo data is effective since it directly turns zero-shot translation into a weakly supervised task. Despite its effectiveness, it brings a lot more training cost during generating data and training on the augmented corpus and the supervised directions' performance is also reported to decrease due to the model capacity bottleneck~\citep{zhang-etal-2020-improving,yang-etal-2021-improving-multilingual}. \citet{rios-etal-2020-subword} finds that instead of regarding all languages as one during the vocabulary building process, language-specific BPE can alleviate the off-target problem, yet it still costs the supervised directions' performance.



\begin{table}[!t]
    \centering
    \includegraphics[width=1\linewidth]{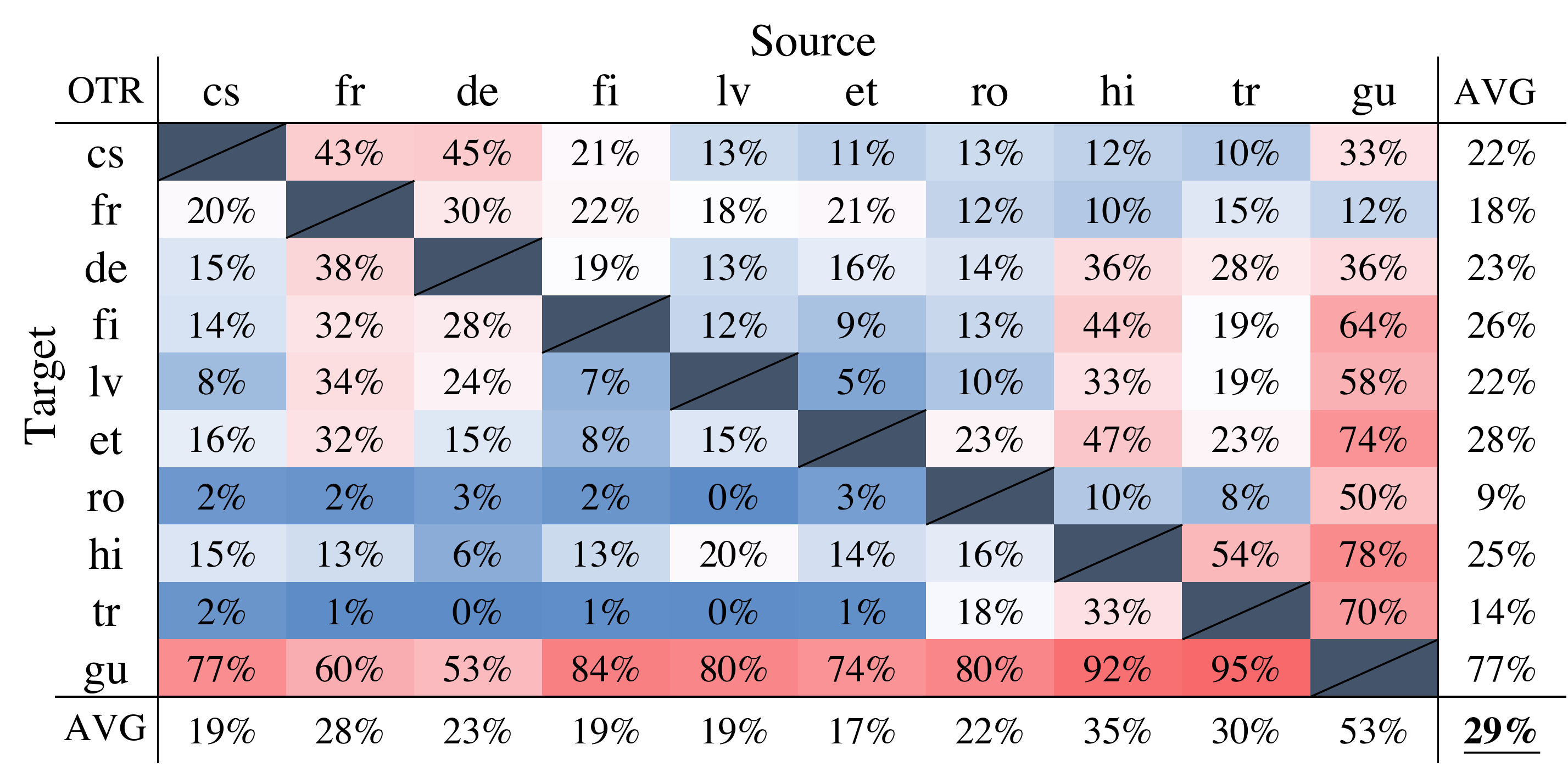}
    \caption{Zero-shot off-target rate of the model with traditional vocab sharing on WMT'10 dataset. High values are in red and low values are in blue. The average OTR of 90 zero-shot directions is about 29\%.}
    \label{tab:otr}
\end{table}

In this work, we perform a comprehensive analysis of the off-target problem, finding that failure in encoding discriminative target language signal will lead to off-target and we also find a strong correlation between off-target rate of certain direction and the lexical similarity between the involved languages. A simple solution by separating the vocabulary of different languages in the decoder can decrease lexical similarity among languages and it proves to improve the zero-shot translation performance. However, it also greatly increases the model size (308M->515M) because a much larger embedding matrix is applied to the decoder.

For a better performance-cost trade-off, we further propose Language-Aware Vocabulary Sharing (LAVS), a novel algorithm to construct the multilingual vocabulary that increases the KL-divergence of token distributions among languages by splitting particular tokens into language-specific ones.

LAVS is simple and effective. It does not introduce any extra training cost and maintains the supervised performance. Our empirical experiments prove that LAVS reduces the off-target rate from 29\% to 8\% and improves the BLEU score by 1.9 points on the average of 90 translation directions. Together with back-translation, the performance can be further improved. LAVS is also effective on larger dataset with more languages such as OPUS-100~\citep{zhang-etal-2020-improving} and we also observe that it can greatly improve the English-to-Many performance (+0.9 BLEU) in the large-scale setting.

\section{Related Work}

  \paragraph{Off-Target Problem in Zero-Shot Translation} Without parallel training data for zero-shot directions, the MNMT model is easily caught up in off-target problem~\citep{ha-etal-2016-toward,aharoni-etal-2019-massively,gu-etal-2019-improved,zhang-etal-2020-improving,rios-etal-2020-subword,wu-etal-2021-language,yang-etal-2021-improving-multilingual} where it ignores the target language signal and translates to a wrong language. Several methods are proposed to eliminate the off-target problem. \citet{zhang-etal-2020-improving,gu-etal-2019-improved} resort different back-translation techniques to generate data for non-English directions. Back-translation method is straight-forward and effective since it provides pseudo data on the zero-shot directions but it brings a lot more additional cost during generating data and training on the augmented corpus. \citet{gu-etal-2019-improved} introduced decoder pretraining to prevent the model from capturing spurious correlations, \citet{wu-etal-2021-language} explored how language tag settings influence zero-shot translation. However, the cause for off-target still remains underexplored.

\paragraph{Vocabulary of Multilingual NMT} Vocabulary building method is essential for Multilingual NMT since it decides how texts from different languages are turned into tokens before feeding to the model. Several word-split methods like Byte-Pair Encoding~\citep{sennrich-etal-2016-neural}, Wordpiece~\citep{Wu2016GooglesNM} and Sentencepiece~\citep{kudo-richardson-2018-sentencepiece}, are proposed to handle rare words using a limited vocab size. In the background of multilingual NMT, most current studies and models \citep{conneau2019unsupervised,Ma2021DeltaLMEP,team2022NoLL} regard all languages as one and learn a shared vocabulary for different languages. \citet{xu-etal-2021-vocabulary} adopted optimal transport to find the vocabulary with most marginal utility. \citet{chen-etal-2022-focus} study the relation between vocabulary sharing and label smoothing for NMT. Closely related to our work, \citet{rios-etal-2020-subword} finds that training with language-specific BPE that allows token overlap can improve the zero-shot scores at the cost of supervised directions' performance and a much larger vocab while our method does not bring any extra cost.

To the best of our knowledge, we are the first to explore how vocabulary similarity of different languages affects off-target in zero-shot MNMT and reveal that solely isolating vocabulary in the decoder can alleviate the off-target problem without involving extra training cost or sacrificing the supervised directions' performance.

\section{Delving into the Off-Target Problem}



\subsection{Multilingual NMT System Description}
\label{section2-1}

We adopt the Transformer-Big~\citep{NIPS2017_attention} model as the baseline model. For multilingual translation, we add a target language identifier <XX> at the beginning of input tokens to combine direction information. We train the model on an English-centric dataset WMT'10~\citep{wmt10}. Zero-shot translation performance is evaluated on Flores-101~\citep{flores1} dataset. We use a public language detector\footnote{https://github.com/Mimino666/langdetect} to identify the sentence-level language and compute the off-target rate (OTR) which denotes the ratio of translation that deviates to wrong languages. Full information about training can be found in Section~\ref{dataset}.

\subsection{Off-Target Statistics Safari}

\begin{figure}[!t]
    \centering
    \includegraphics[width=1.0\linewidth]{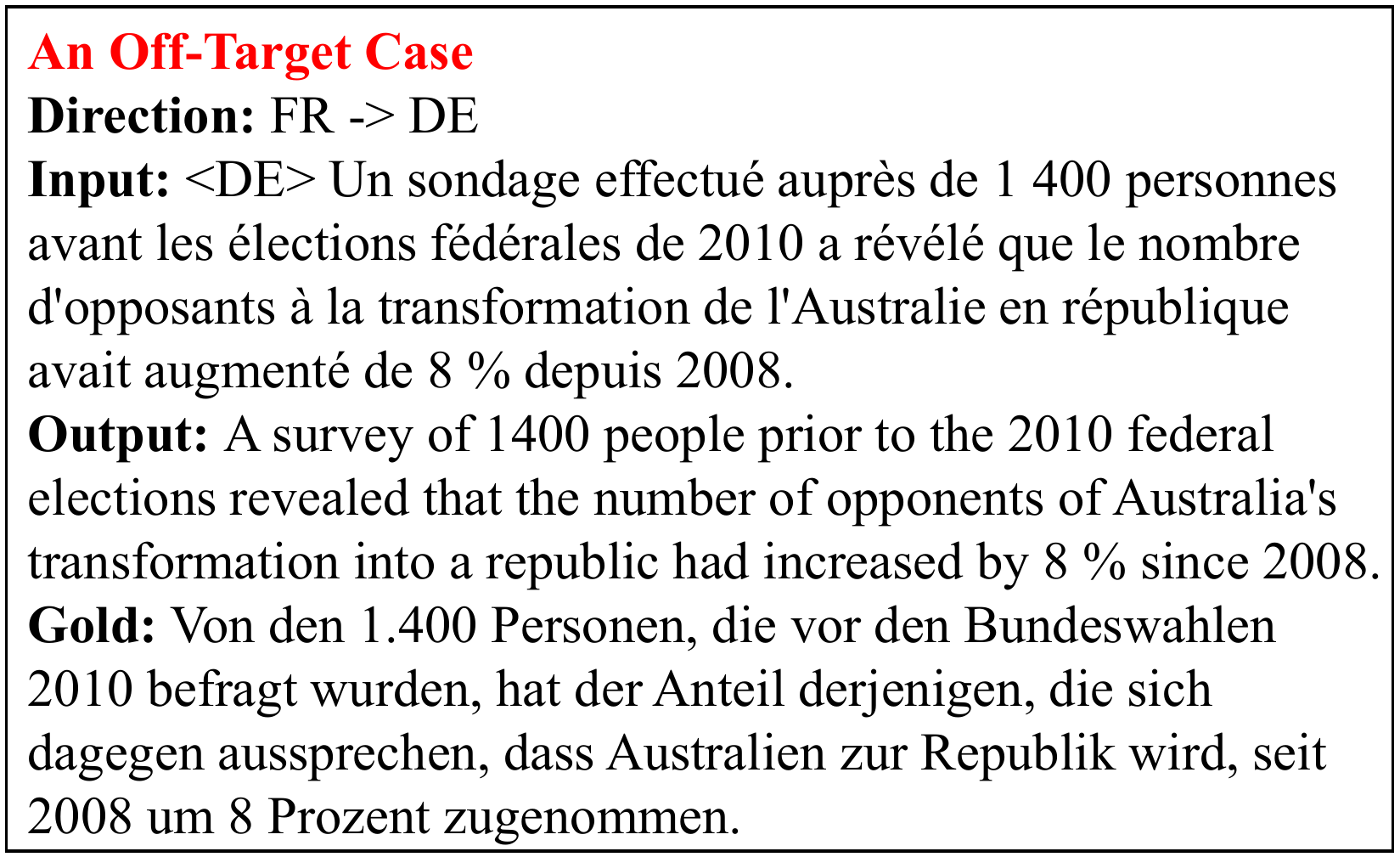}
    \caption{A real Off-Target case observed in our multilingual NMT system. In this case, the output is literally English while the real target is German. }
    \label{fig:safari}
\end{figure}

\paragraph{Off-Target Rate Differs in Directions} 

We first train the multilingual NMT model in 10 EN-X directions and 10 inverse directions from WMT'10 simultaneously. Then we test the model on 90 X-Y zero-shot directions using semantic parallel sentences from the previous 10 languages provided by Flores-101. We compute the off-target rate of all directions and list the result in Table~\ref{tab:otr}.

In addition to the individual score, we next split the languages into High (cs, fr, de, fi), Mid (lv, et), and Low (ro, tr, hi, gu) resources according to data abundance degree. Then we compute the average OTR of High-to-High, High-to-Low, Low-to-High, and Low-to-Low directions and rank the result. The ranked result is: Low-to-Low (50.28\%) > High-to-High (27.16\%) > Low-to-High (23.18\%) > High-to-Low (20.78\%). Based on the observation, we can see that language with the lowest resource (gu) contributes to a large portion of off-target cases. This is reasonable since the model might not be familiar with the language identifier <GU> and the same situation goes for Low-to-Low translations. 

However, it is surprising to see that translations between high-resource languages suffer from more severe off-target than those directions involving one low-resource language. There seem to be other factors influencing the off-target phenomena.

In other words, if data imbalance is not the key factor for off-targets between high-resource languages, what are the real reasons and possible solutions? To answer these questions, we need to delve deeper into the real off-target cases.

\paragraph{The Major Symptom of Off-Target} 

When the model encounters an off-target issue, a natural question is which language the model most possibly deviates to. We find that among different directions, a majority (77\%) of the off-target cases are wrongly translated to English, which is the centric language in the dataset. A small part (15\%) of cases copy the the input sentence as output. Our observation also agrees with the findings of \citet{zhang-etal-2020-improving}. It raises our interest that why most off-target cases deviate to English. 

    



\subsection{Failing in Encoding Discriminative Target Language Signal Leads to Off-Target}
\label{2.4}

Considering the encoder-decoder structure of the model, we hypothesize that: 

\textit{The encoder fails to encode discriminative target language information to the hidden representations before passing to the decoder. }

\begin{figure}[t]
    \centering
    \includegraphics[width=1.0\linewidth]{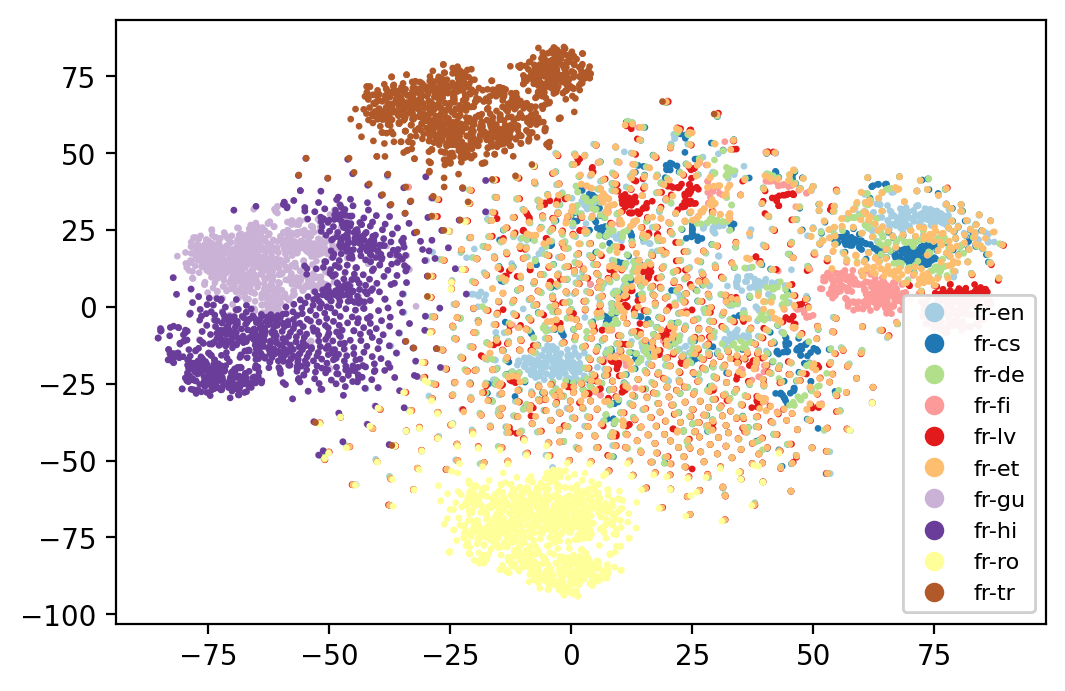}
    \caption{Encoder pooled output visualization using TSNE for French-to-Many translations. The input French sentences are the same for all directions. Note that there are only French sentences in the encoder side.}
    \label{fig:PCA_a}
\end{figure}

To test the hypothesis, we start by analyzing the output of the trained transformer's encoder:

1) We choose French as the source language and conduct a French-to-Many translation (including all languages in WMT'10) on Flores-101. 

2) We collect all the pooled encoder output representations of the French-to-Many translation and project them to 2D space using TSNE. The visualization result is shown in Figure~\ref{fig:PCA_a}.

The visualization result justifies our hypothesis. We can tell from the distribution that only representations belonging to ``fr-tr'' and ``fr-ro'' directions have tight cluster structures with boundaries. \textit{The representations from high/mid-resource language pairs are completely in chaos and they are also mixed with fr-en representations.}  And those languages generally have a higher off-target rate in French-to-Many Translation according to Table~\ref{tab:otr}. 

The decoder cannot distinguish the target language signal from the encoder's output when it receives representations from the ``chaos'' area. Moreover, during the training process, the decoder generates English far more frequently than other languages and it allocates a higher prior for English.

Passing hidden representation similar to English one will possibly confuse the decoder to generate English no matter what the given target language is. It could explain why most off-target cases deviate to English. The decoder struggles to tell the correct direction from the encoder's output.

Now we have a key clue for the off-target issue. The left question is what causes the degradation of target language signal in some directions and whether we can make the representations of different target languages more discriminative to eliminate the off-target cases. 


    

\subsection{Language Proximity Correlates with Zero-Shot Off-Target Rate }

To explore how off-target occurs differently in different language pairs, we conduct experiments using a balanced subset of WMT'10 dataset where we hope to preclude the influence of data size. We randomly sampled 500k sentences from different directions to form a balanced training set and remove the directions(hi, tr and gu) that do not have enough sentences.

\paragraph{Language Proximity is an Important Characteristic of Translation Direction} 




Our motivation is intuitive that if two languages are rather close, the probability distribution of different n-grams in the two languages' tokenized corpus should be nearly identical. Considering a large number of different n-grams in the corpus, we only consider 1-grams to compute the distribution. We call the result ``Token Distribution''.

We use Kullback–Leibler divergence from Token Distribution of Language B to Language A to reflect the degree of difficulty if we hope to encode sentence from B using A, which can also be interpreted as ``Lexical Similarity''.


\begin{equation}
D_{\mathrm{KL}}(A \| B)=\sum_{x \in \mathcal{V}} A(x) \log \left(\frac{A(x)}{B(x)}\right)
\end{equation}

where $\mathcal{V}$ denotes the shared vocabulary, $A(x)$ is the probability of token $x$ in language $A$. To avoid zero probability during computing Token Distribution, we add 1 to the frequency of all tokens in the vocabulary as a smoothing factor.

\paragraph{Lexical Similarity is related to Off-Target Rate} We compute the KL divergence between language pairs with the training data. After training on the balanced dataset, the zero-shot translation is conducted on the Flores-101 dataset. We visualize the result of the top-3 languages(fr,cs,de) with most resources in WMT'10 dataset for analysis.

\begin{figure}[t]
    \centering
    \includegraphics[width=1.0\linewidth]{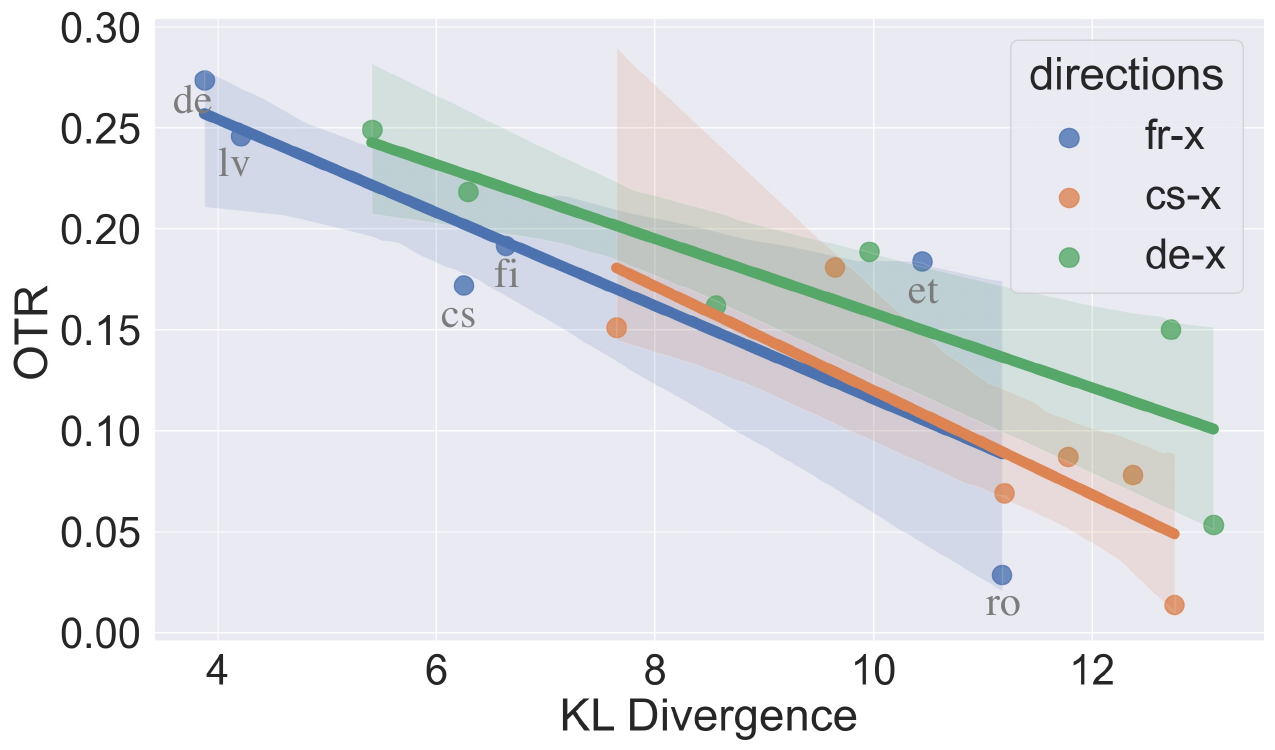}
    \caption{Scatter plot of off-target rate and KL-divergence for different language pairs. We draw the linear  regression result with 95\% confidence interval.}
    \label{fig:scatter}
\end{figure}

As shown in Figure~\ref{fig:scatter}, we can observe from the statistics that language proximity is highly related to the off-target rate. The Pearson correlation coefficients between the off-target rate and the KL-Divergence from target to source of the three x-to-many translations are -0.75±0.02, -0.9±0.03 and -0.92±0.03.  The average Pearson correlation of all x-to-many directions is -0.77±0.11. It indicates that language pair which has higher lexical similarity from target to source may have a higher chance to encounter off-target than those language pairs which has less similar languages. 

\subsection{Shared Tokens in the Decoder Might Bias the Zero-Shot Translation Direction} 
\label{sec3-5}

Previous section shows a correlation between the lexical similarity and off-target rate within certain language pair. We are more interested in whether the lexical similarity will cause the representation degradation in Figure~\ref{fig:PCA_a}, which further causes off-target. In fact, larger lexical similarity suggests more shared tokens between languages and will let the decoder output more overlapped tokens during supervised training. \textbf{The token overlap for different target in output space is harmful for zero-shot translation.} During training, the decoder might not be aware of the language it's generating directly from the output token because of the existence of shared tokens. In other words, the relation between target language and output tokens is weakened because of the shared tokens among different target languages, which might cause representation degradation in the encoder and further lead to off-target in zero-shot test.




\subsection{Separating Vocab of Different Languages is Effective yet Expensive }

Based on the previous discussion, we now have an idea that maybe we can ease the off-target problem by decreasing the lexical similarity among languages, i.e. decreasing the shared tokens. 

When building the vocab for multilingual NMT model, most work regard all languages as one and learn a unified tokenization model. We argue that this leads to low divergence of token distribution since many sub-words are shared across languages. 

There is an easy method to decrease the shared tokens without changing the tokenization. We can separate the vocab of different languages as shown in Figure~\ref{fig:complete_split} from Appendix. Under such condition, no two languages share the same token.


\begin{table}[t]
    \centering
       \resizebox{0.45\textwidth}{!}{
    \begin{tabular}{lccc}
\toprule
Method & Size & OTR &BLEU \\
\midrule 
Vocab Sharing&308M & 29\% & 10.2 \\
Separate Vocab (Dec)&515M&\textbf{5\%}  & \textbf{12.4} \\
Separate Vocab (Enc,Dec)&722M& 84\%   & 2.1\\

\bottomrule
\end{tabular}}
    
    \caption{Average zero-shot result for models with different vocab. (Dec) means only the decoder uses the separate vocab. (Enc,Dec) means both the encoder and the decoder use the separate vocab.}
    \label{tab:easy_solution}
\end{table}

\begin{figure}[t]
    \centering
    \includegraphics[width=1.0\linewidth]{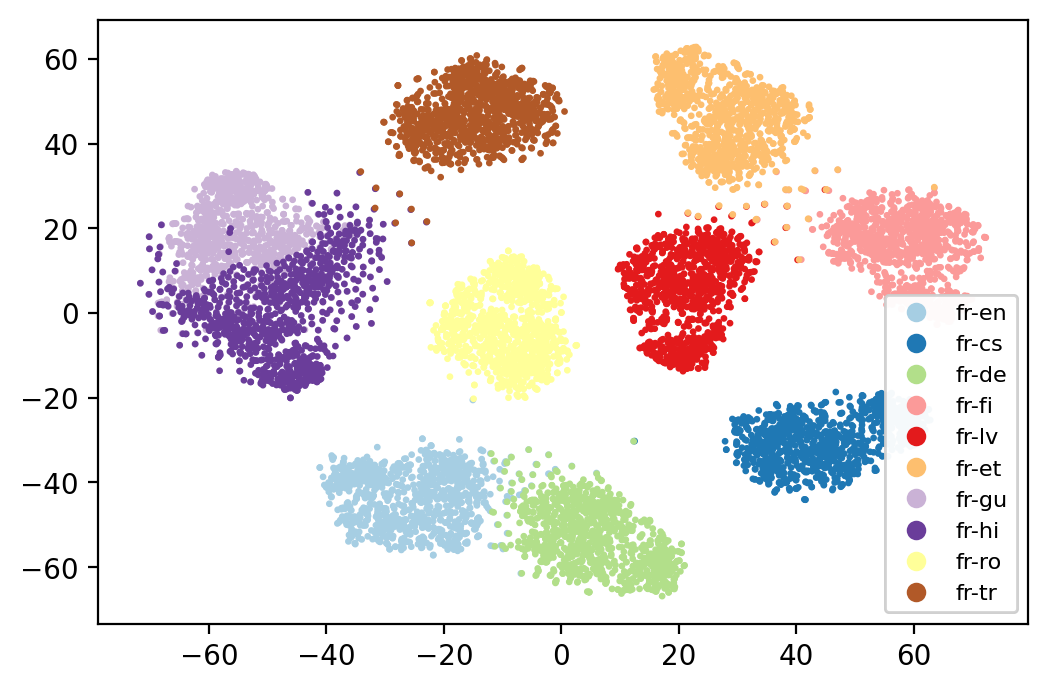}
    \caption{Encoder pooled output visualization using TSNE for French-to-Many translation using separate vocab. The result is comparable to Figure~\ref{fig:PCA_a}, which shows result with shared vocab.}
    \label{fig:PCA_b}
\end{figure}

 As shown in Table~\ref{tab:easy_solution}, with separate decoder vocab the average off-target rate in 90 directions is reduced from 29\% to 5\% and the BLEU score is raised from 10.2 to 12.4. We conduct the same probing experiment on encoder representation with the original WMT'10 dataset. As shown in Figure~\ref{fig:PCA_b}, representations for different target are divided. The ``chaos'' area does not exist anymore. 
 
 We also train the model with separated encoder\&decoder vocab and finds it suffers from worse zero-shot performance compared to baseline. This also agrees to \citet{rios-etal-2020-subword}'s findings.
 
 We think that without any vocabulary sharing among languages, the model will learn a wrong correlation between input language and output language and ignore the target language identifier during the English-centric training process.

The experiment result justifies our assumption in Section~\ref{sec3-5} that the shared tokens in the decoder will lead to the representation problem. Though achieving great improvement by isolating all vocabulary, it is much more parameter-consuming. In fact, in our experiment, the number of parameters increases from 308M to 515M. 


\section{Language-Aware Vocabulary Sharing}

\subsection{Adding Language-Specific Tokens}
\begin{figure}[h]
    \centering
    \includegraphics[width=0.8\linewidth]{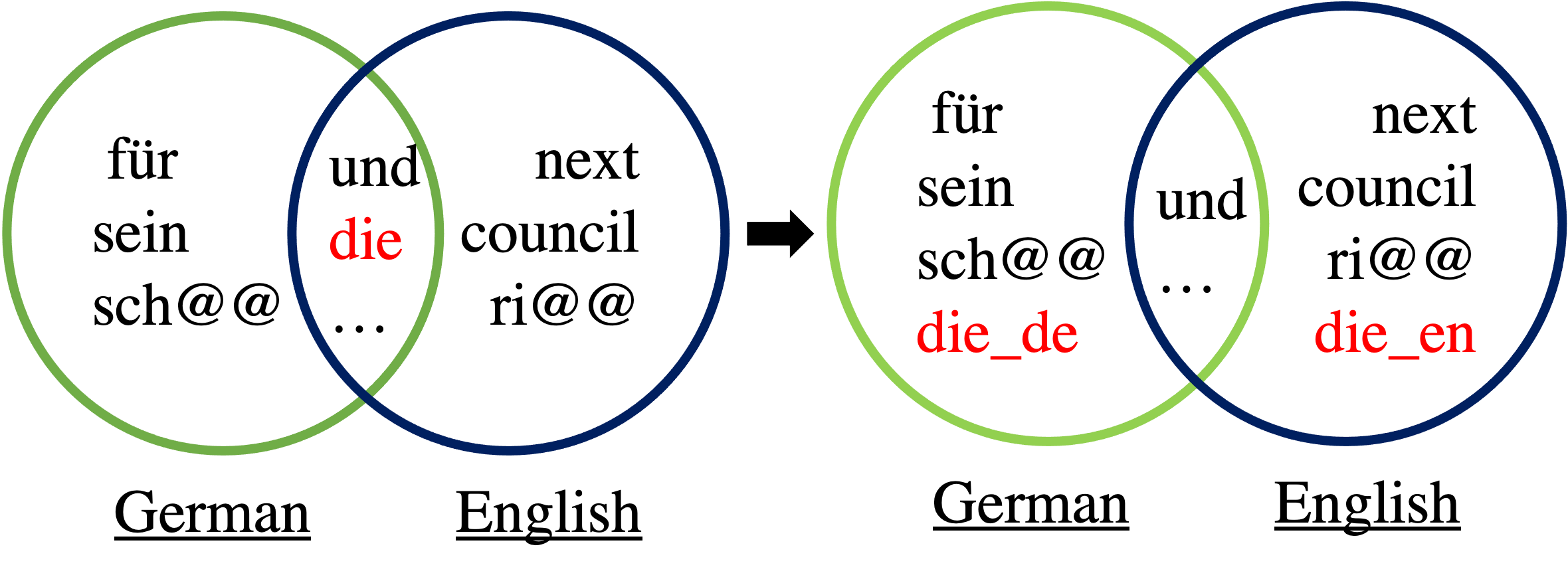}
    \caption{Illustration of LAVS. Tokens with higher shared frequency are split into language-specific ones.}
    \label{fig:vocab_split}
\end{figure}

\begin{algorithm}[b]
\caption{Language-Aware Vocabulary Sharing} 
  \small {\hspace*{0.02in} {\bf Input:} 
Shared vocabulary set $V'$, language list $L$, language's token distributions $P$ and the number of extra language-specific tokens $N$.  \\}
\hspace*{0.02in} {\bf Output:} 
  $V_{out}$ is the output vocabulary set. 
\begin{algorithmic}[1]
\State MaxFreqs = PriorQueue(length=$N$)\Comment{ queue that ranks the input elements E from high to low based on E[0].}
\For{$i$ in $V'$} 
\For{$m$ in $L$, $n$ in $L$}

\If{$m<n$}

\State freq = min($P_m^{V'}(i)$,$P_n^{V'}(i)$) 
\State MaxFreqs.add([freq,$m$,$n$,$i$])

\EndIf
\EndFor
\EndFor
\State  $V_{out}$ = $V'$
\For{T in MaxFreqs}
\State $m, n, i$ = T[1], T[2], T[3]
\State $V_{out}$ = $V_{out} \cup (V'[i],{L[m]}) \cup (V'[i],{L[n])}$
\EndFor
\State \Return $V_{out}$
\end{algorithmic}
\label{LAVS}
\end{algorithm}
Based on previous observation, lexical similarity will cause the representation degradation problem and further lead to off-target. Thus, our goal is to decrease the lexical similarity. We can achieve it without changing the original tokenizer by splitting the shared tokens into language-specific ones.

As shown in Figure~\ref{fig:vocab_split}, instead of splitting all shared tokens, we can choose specific tokens to split. After decoding, we could simply remove all language-specific tags to restore the literal output sentence. By adding language-specific tokens, the number of shared tokens between different languages decreases and makes the token distribution more different thus increasing the KL Divergence.

\subsection{Optimization Goal}

Given original vocab set $V'$ and language list $L$, we aim at creating new vocab $V$ to maximize the average KL divergence within each language pair under the new vocabulary with the restriction of adding $N$ new language-specific tokens. Thus, our objective becomes:

\begin{equation}
\begin{aligned}
V^*=&\mathop{\arg\max}\limits_{V} \frac{1}{|L|^2} \sum_{m\in L}\sum_{n\in L} D_{KL}(P_m^V || P_n^V) \\
&s.t. \quad  V'\subseteq V, \quad|V|-|V'| = N 
\end{aligned}
\end{equation}
where $P_m^V$ denotes the $m$-th language's token distribution on vocabulary $V$, add-one smoothing is applied to avoid zero probability. It is a combinatorial optimization problem. The searching space of V has an astronomical size of  $C_{|V'|\cdot|L|}^N$.




\begin{figure}[t]
    \centering
    \includegraphics[width=1.0\linewidth]{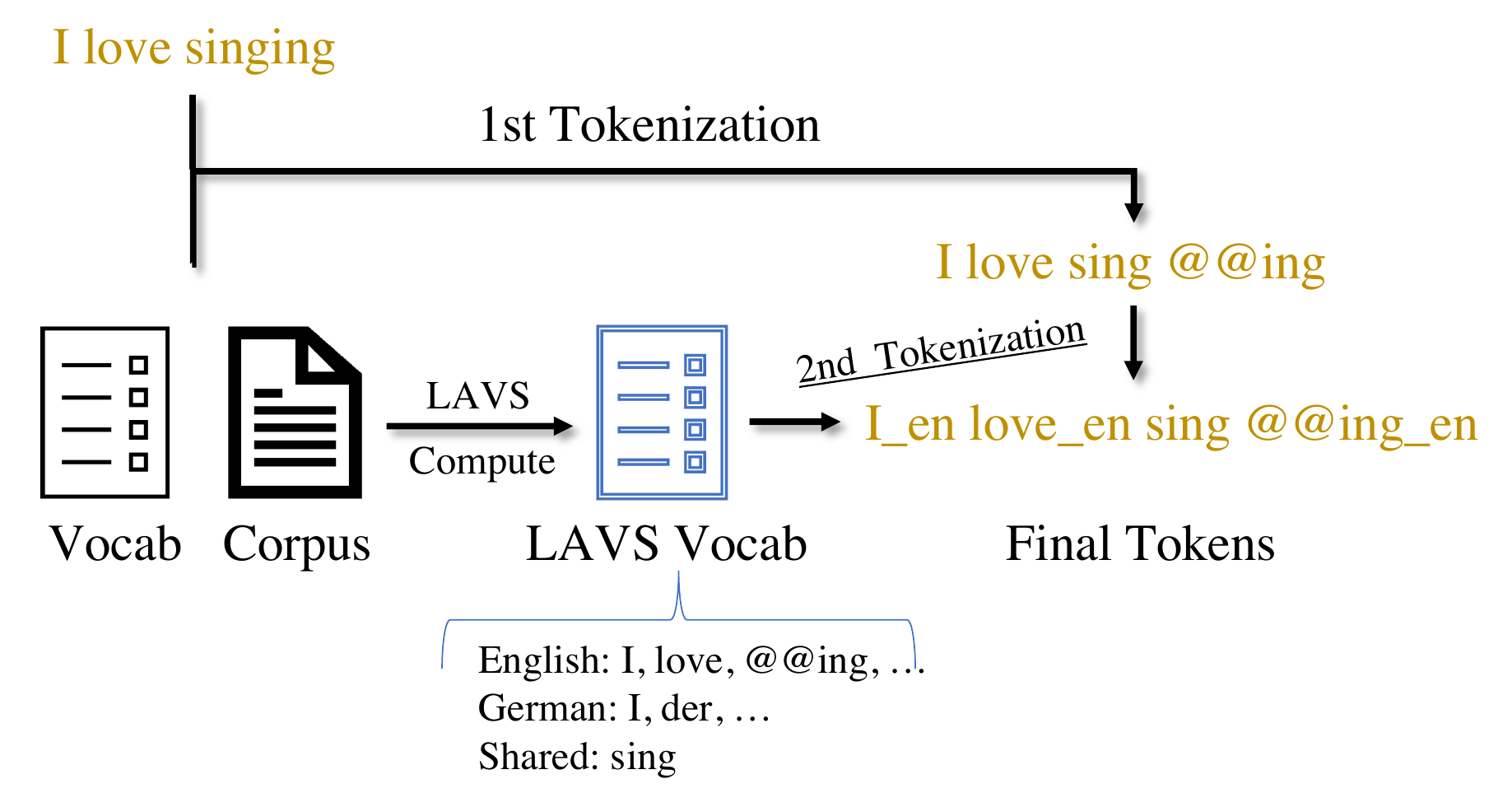}
    
    \includegraphics[width=1.0\linewidth]{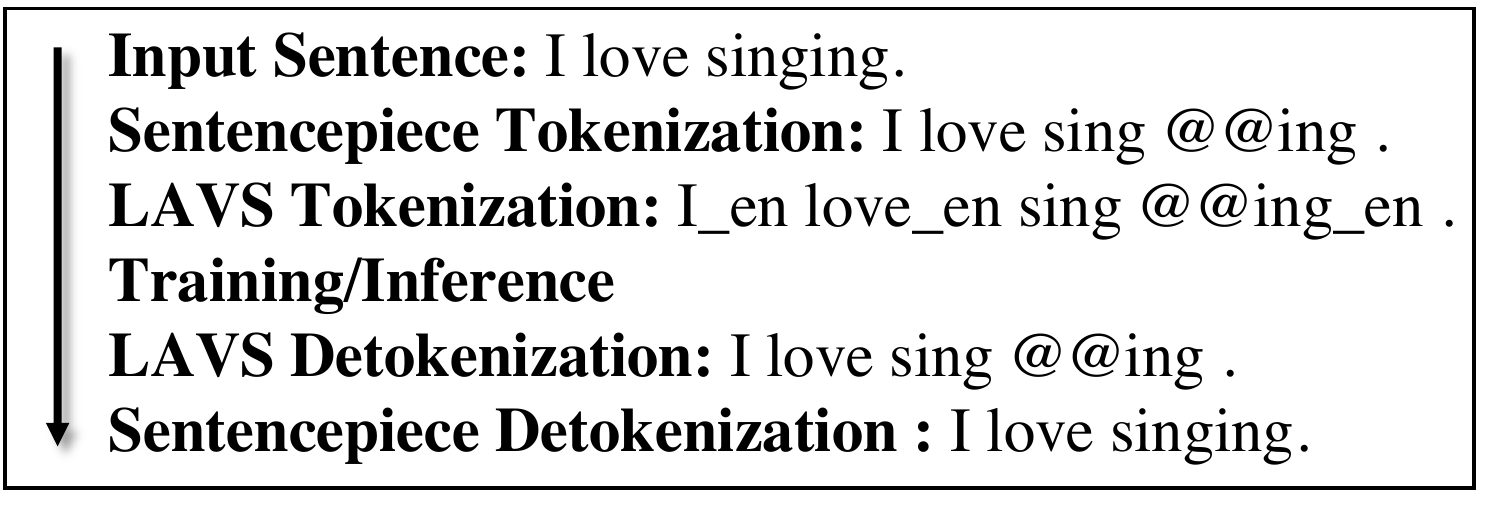}
    \caption{Illustration of tokenization and detokenization process with Language-Aware Vocabulary Sharing. }
    \label{fig:2token}
\end{figure}

\subsection{Greedy Selection Algorithm that Maximizes Divergence Increment}

\begin{table*}[t]
    \centering
    \resizebox{1.0\textwidth}{!}{
    \begin{tabular}{lccccccccccccccc}
    \toprule
        
        \multirow{2}{*}{Method} &
        \multirow{2}{*}{Size}&\multicolumn{5}{c}{Zero-Shot Off-Target Rate} & \multicolumn{7}{c}{BLEU Score} \\
                \cmidrule(r){3-7} \cmidrule(r){8-14}
                
        ~ & ~&x-y &H-H& L-L & H-L &L-H &x-y &H-H& L-L & H-L &L-H &en-x & x-en

        \\ \midrule
        Vocab Sharing & 308M &29\%  &27\% & 50\% & 21\%& 23\%&10.2&11.26&5.03 &9.18 &9.95 &24.8 & 30.2  \\
        Separate Vocab (Dec) & 515M  &\textbf{5\%}& 4\% & 19\% & \textbf{1\%} &\textbf{1\%} & 12.4 &14.69&6.54&\textbf{10.10}&\textbf{12.22} & 24.6 & \textbf{30.5}  \\
        LAVS (Enc, Dec) & 308M &12\% & \textbf{3\%}&33\% & 13\%&6\%&\textbf{12.5}&\textbf{15.90} &6.26 &9.91 &12.14 &24.8 &30.3  \\
        LAVS (Dec) &308M   &8\% & 13\% & \textbf{14\%}  & 3\% &4\%& 12.1 &13.33&\textbf{7.81}&9.80&12.01 &\textbf{24.9}&30.3\\

        \bottomrule
    \end{tabular}}
    \caption{Overall performance comparison. x-y denotes all zero-shot directions. H and L denotes High/Low-resources.  All evaluation are done with Flores-101 dataset. (Dec) suggests vocab only changes in decoder and (Enc, Dec) suggests changing in both encoder and decoder. LAVS outperforms baseline in zero-shot setting on both BLEU and OTR by a large margin while maintaining the en-x and x-en performance.}
    \label{tab:4}
\end{table*}

\begin{table*}[t]
    
    \centering
    \resizebox{1.0\textwidth}{!}{
    \begin{tabular}{clcccccccccc}
    \toprule
        \makebox[0.05\textwidth][c]{Metric} &Method & cs-x & fr-x & de-x & fi-x & lv-x & et-x & ro-x & hi-x & tr-x & gu-x \\ \midrule
        
        \multirow{3}{*}{\rotatebox[origin=c]{360}{{OTR }}}    
        
        ~&Vocab Sharing &18.8\% & 28.3\% & 22.6\% &19.5\% &19.2\% &17.1\% & 22.0\% & 35.2\% & 30.1\% & 52.8\% \\
        ~&LAVS(Dec)  & \textbf{4.2\%} & \textbf{14.4\%} & \textbf{11.5\%} & \textbf{6.2\%} & \textbf{3.7\%} & \textbf{4.7\%} & \textbf{2.9\%} & \textbf{9.7\%} & \textbf{10.2\%} & \textbf{6.1\%} \\
        ~&$\Delta\downarrow$ & -14.6\% & -13.9\% & -11.1\% & -13.3\% & -15.5\% & -12.4\% & -19.1\% & -25.5\% & -19.9\% & -46.7\% \\
        
        \midrule
        \midrule

        \multirow{3}{*}{\rotatebox[origin=c]{360}{{BLEU}}}    
        
        ~&Vocab Sharing&10.9&10.5&11.3&9.0&9.4&10.0&11.7&6.9&7.3&4.7\\
        ~&LAVS(Dec) &\textbf{12.0} & \textbf{12.0} & \textbf{12.2} & \textbf{9.6} & \textbf{10.9} & \textbf{11.0} & \textbf{14.0} & \textbf{9.3} & \textbf{9.1} & \textbf{8.4}  \\
       ~&$\Delta\uparrow$& +1.1  &  +1.5  &  +0.9  &  +0.6  &  +1.5  &  +1.0  &  +2.3  &  +2.4  &  +1.8  &  +3.7 \\
       
        \midrule
        \midrule
        \multirow{3}{*}{\rotatebox[origin=c]{360}{{BERT Score}}}  
        ~&Vocab Sharing  &0.781 & 0.808 & 0.787 & 0.766 & 0.783 & 0.774 & 0.791 & 0.771 & 0.643 & 0.677\\
        ~&LAVS(Dec) &\textbf{0.799} & \textbf{0.829} & \textbf{0.806} & \textbf{0.786} & \textbf{0.790} & \textbf{0.798} & \textbf{0.796} & \textbf{0.777} & \textbf{0.660} & \textbf{0.713} \\
        ~&$\Delta\uparrow$ &0.018 & 0.021 & 0.019 & 0.020 & 0.007 & 0.024 & 0.005 & 0.006 & 0.017 & 0.036 \\

        \bottomrule
        ~ & ~ & ~ & ~ & ~ & ~ & ~ & ~ & ~ & ~ & ~ \\ \toprule
        Metric& Method & x-cs & x-fr & x-de & x-fi & x-lv & x-et & x-ro & x-hi & x-tr & x-gu \\ \midrule

        \multirow{3}{*}{\rotatebox[origin=c]{360}{{OTR }}}  
        ~&Vocab Sharing&22.4\%&17.8\%&23.9\%&26.0\%&21.9\%&28.1\%&8.9\%&25.4\%&14.0\%&77.0\%\\
        ~&LAVS(Dec) & \textbf{8.7\%} & \textbf{5.9\%} & \textbf{6.6\%} & \textbf{9.2\%} & \textbf{8.4\%} & \textbf{7.8\%} & \textbf{3.0\%} & \textbf{1.7\%} & \textbf{7.0\%} & \textbf{15.4\%} \\
        ~&$\Delta\downarrow$ & -13.7\% & -11.9\% & -17.3\% & -16.8\% & -13.5\% & -20.3\% & -5.9\% & -23.7\% & -7.0\% & -61.6\% \\
        
        \midrule
        \midrule
        \multirow{3}{*}{\rotatebox[origin=c]{360}{{BLEU }}}  
        ~&Vocab Sharing&11.0&17.9&13.2&8.3&12.2&9.9&14.0&8.3&8.8&3.3\\
        ~&LAVS(Dec) & \textbf{12.5} & \textbf{20.1} & \textbf{15.7} & \textbf{9.4} & \textbf{13.3} & \textbf{11.7} & \textbf{14.2} & \textbf{9.9} & \textbf{9.0} & \textbf{6.7} \\
        ~&$\Delta\uparrow$ & +1.5  &  +2.2  &  +2.5  &  +1.1  &  +1.1  &  +1.8  &  +0.2  &  +1.6  &  +0.2  &  +3.4 \\
        
        \midrule
        \midrule
        \multirow{3}{*}{\rotatebox[origin=c]{360}{{BERT Score}}}  
        ~&Vocab Sharing &0.772 & 0.776 & 0.781 & 0.749 & 0.757 & 0.759 & 0.771 & 0.743 & 0.750 & 0.723 \\
        ~&LAVS(Dec) & \textbf{0.791} & \textbf{0.799} & \textbf{0.796} & \textbf{0.770} & \textbf{0.777} & \textbf{0.774} & \textbf{0.797} & \textbf{0.756} & \textbf{0.768} & \textbf{0.726} \\
        ~&$\Delta\uparrow$  & 0.019 & 0.023 & 0.015 & 0.021 & 0.020 & 0.015 & 0.026 & 0.013 & 0.018 & 0.003 
        
        \\ \bottomrule
    \end{tabular}}
    \caption{The zero-shot translation performance (Off-Target Rate, BLEU and BERT-Score) on average x-to-many and many-to-x directions using LAVS (Dec) compared to baseline. }

    \label{tab:3}
\end{table*}

Based on the previous discussion, we propose the Language-Aware Vocabulary Sharing algorithm as listed in Algorithm~\ref{LAVS} to add language-specific tokens. Intuitively, LAVS algorithm prefers to split those shared tokens that have high frequency among different languages, which directly reduces the appearance of shared tokens in the decoder to the maximum extent.

First, we adopt a prior queue to keep the token candidates. Second, for each token in the shared vocabulary, we compute the shared token frequency in each language pair and add the (frequency, languageA, languageB, token) tuple to the queue. Last, since the queue ranks the elements by frequency, we create language-specific tokens for the top $N$ tuples and return the new vocab. We give more details about the algorithm in Appendix~\ref{app:alg}.  
 
 The whole tokenization process with LAVS is illustrated in Figure~\ref{fig:2token}. In practice, given an original shared vocab with $M$ tokens, we can always first learn a vocab with $M-N$ tokens and conduct LAVS to add $N$ language-specific tokens to maintain the vocab size $M$ unchanged.


\section{Experiments}

\subsection{Datasets}
\label{dataset}

Following \citet{wang-etal-2020-multi}, we collect WMT'10 datasets for training. The devtest split of Flores-101 is used to conduct evaluation. Full information of datasets is in Appendix~\ref{app:dataset}.

\subsection{Vocabulary Building}

\paragraph{Vocab Sharing} We adopt Sentencepiece~\citep{kudo-richardson-2018-sentencepiece} as the tokenization model. We randomly sample 10M examples from the training corpus with a temperature of 5\citep{wild} on different directions and learn a shared vocabulary of 64k tokens. 

\paragraph{Separate Vocab} Based on the sharing vocab of the baseline model, we separate the vocab of each language forming a 266k vocab.

\paragraph{LAVS} We first learn a 54k vocabulary using the same method as the baseline model's and add 10k language-specific tokens using LAVS.

\subsection{Training Details of MNMT}

\paragraph{Architecture} We use the Transformer-big model~\citep{NIPS2017_attention} implemented by fairseq~\citep{ott2019fairseq} with $d_{model}=1024$, $d_{hidden}=4096$, $n_{heads}=16$, $n_{layers}=6$. We add a target language identifier <XX> at the beginning of input tokens to indicate the translation directions as suggested by~\citet{wu-etal-2021-language}.

\paragraph{Optimization} We train the models using Adam~\citep{Kingma2015AdamAM}, with a total batch size of 524,288 tokens for 100k steps in all experiments on 8 Tesla V100 GPUs. The sampling temperature, learning rate and warmup steps are set to 5, 3e-4 and 4000.

\paragraph{Back-Translation} Back-Translation method is effective in improving zero-shot performance by adding pseudo parallel data generated by the model~\citep{gu-etal-2019-improved,zhang-etal-2020-improving}. For simplicity, we apply off-line back-translation to both the baseline and LAVS. With the trained model, we sample 100k English sentences and translate them to other 10 languages, which creates 100k parallel data for every zero-shot language pair and results in a fully-connected corpus of 9M sentence pairs. We add the generated data to the training set and train the model for another 100k steps.

\paragraph{Evaluation} We report detokenized BLEU using sacrebleu\footnote{nrefs:1|case:mixed|eff:no|tok:13a|smooth:exp|version:2.1.0}. We also report the Off-Target rate with language detector\footnote{https://github.com/Mimino666/langdetect} and conduct model-based evaluation using Bert-Score\footnote{https://github.com/Tiiiger/bert\_score} \citep{bert-score}.

\subsection{Results}

\paragraph{LAVS improves zero-shot translation by a large margin.}
Table~\ref{tab:4} and \ref{tab:3} list the overall results on both zero-shot and supervised directions. According to Table~\ref{tab:4}, we can see that LAVS improves all the x-to-many and many-to-x directions with a maximum average improvement of -61.6\% OTR, +3.7 BLEU and +0.036 Bert-Score compared to the baseline vocab. It gains an average of -21\% OTR, +1.9 BLEU and +0.02 Bert-Score improvement on 81 zero-shot directions. Compared with the Separate Vocab (Dec) method which also leads to significant improvement in x-y directions, LAVS does not increase any model size.

\paragraph{LAVS with Back-Translation further improves the zero-shot performance.}

\begin{table}[t]
    \centering
       \resizebox{0.47\textwidth}{!}{
    \begin{tabular}{lcccccc}
\toprule
Data & OTR & x-y & en-x & x-en & Extra Cost \\

\midrule
Vocab Sharing & {29\%}  & {10.2}  & {24.8}  & {30.2}  & -  \\
\quad \quad \quad + B.T. &{1\%}   & {16.4}  & {23.4} &{30.0}  & 24 GPU Days\\
LAVS (Dec) &8\% &12.1 &\textbf{24.9} &30.3 & 0\\
\quad \quad \quad + B.T. &\textbf{0\%} &\textbf{16.8} &23.7 &\textbf{30.4} & 24 GPU Days\\

\bottomrule
\end{tabular}}
    \caption{Results with Back-Translation.}
    \label{tab:bt}
\end{table}

As shown in Table~\ref{tab:bt}, as expected, our back-translation method can improve the zero-shot performance by a large margin. Under such setting, LAVS also outperforms Vocab Sharing by 0.4 average BLEU score on zero-shot directions. 

We also observe performance degradation in English-to-Many directions for both models comparing to not using back-translation, which also agrees to the result of~\citet{zhang-etal-2020-improving,rios-etal-2020-subword}. We think a possible reason is that the English-to-Many performance will be interfered with the increase of translation tasks. Back Translation also brings much extra cost. The total training time for the model with Back-Translation is almost twice as long as the model with vanilla training. Only applying LAVS brings no extra training cost and does not influence the supervised performance.

\section{Discussion}

\subsection{How does LAVS calibrate the direction?}
\begin{figure}[t]
    \centering
    \includegraphics[width=1.0\linewidth]{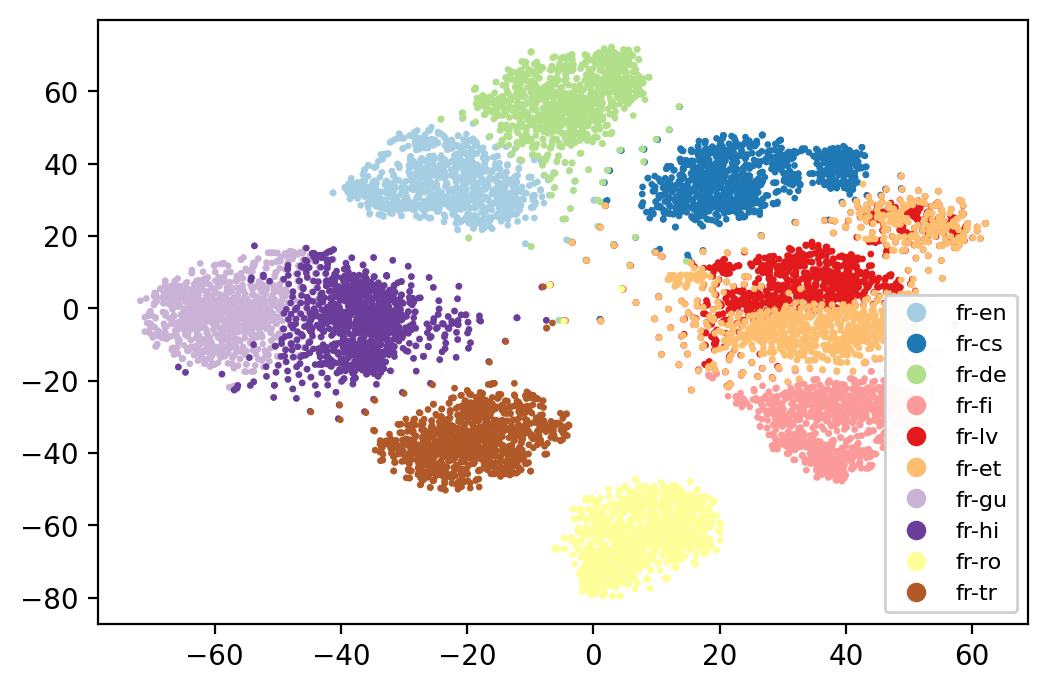}
    \caption{The encoder-pooled representations learned by multilingual NMT with LAVS on fr-x directions.}
    \label{fig:lavs-tsne}
\end{figure}

We visualize the encoder-pooled representations for model with LAVS(dec) in Figure~\ref{fig:lavs-tsne}. The representations' distribution is similar to Figure~\ref{fig:PCA_b} where representations for different target are almost divided, suggesting that LAVS work similarly to separating all the vocabulary for different languages. We also give a case study as shown in Section~\ref{sec:case}.

We further visualize the language identifiers' hidden output during among high-resource languages and compare the results of the original Vocabulary Sharing and LAVS. As shown in Figure~\ref{fig:token_XX} from Appendix, it turns out that LAVS encodes more discriminative target language information into the <XX> token's hidden output.


\subsection{Case Study}
\label{sec:case}

\begin{figure}[h]
    \centering
    \includegraphics[width=1\linewidth]{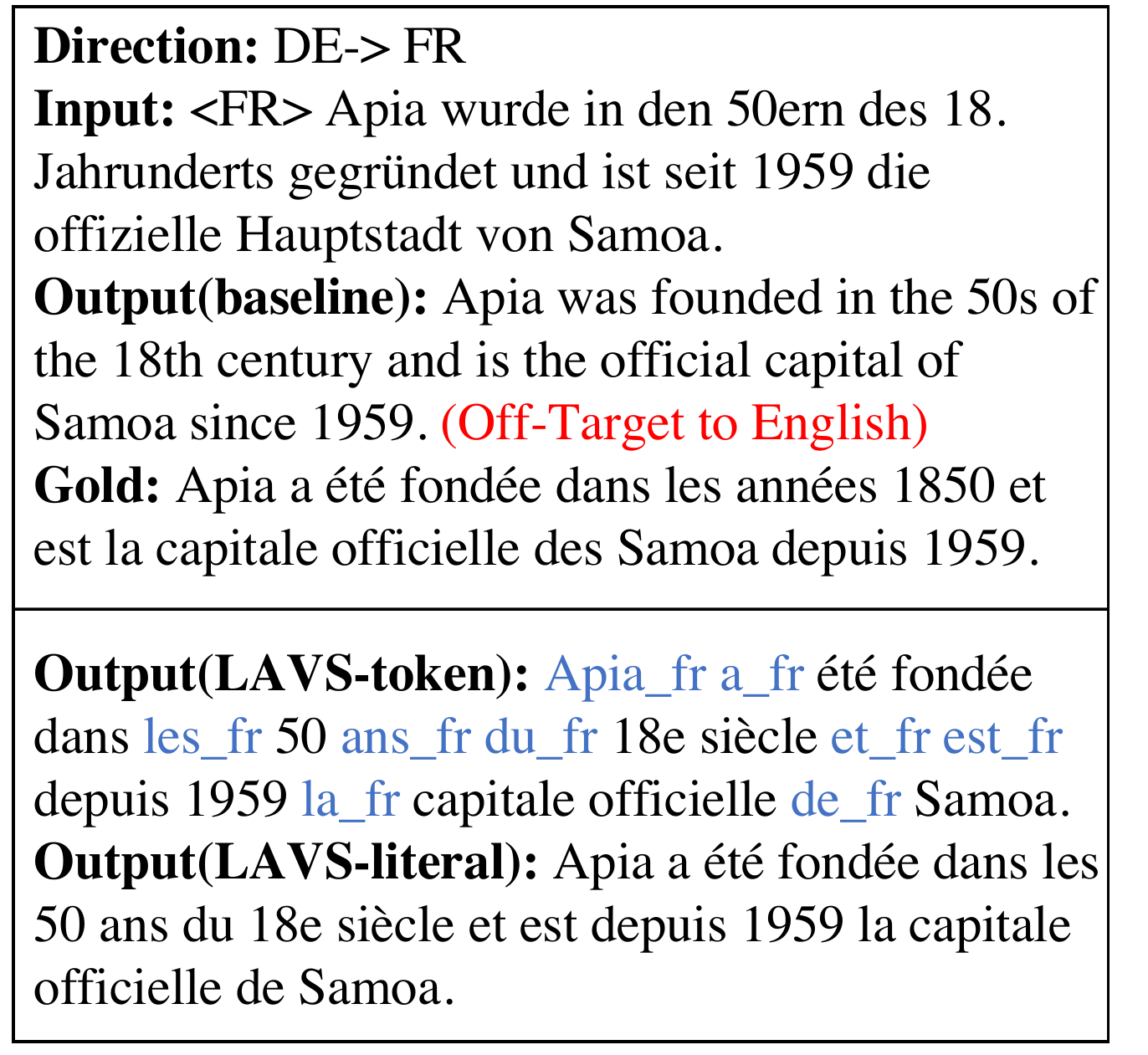}
    \caption{Case study of DE->FR zero-shot translation. The baseline model off-target to English. Tokens in blue belong to language-specific tokens.}
    \label{fig:case_eng}
\end{figure}

We compare different model's outputs as shown in Figure~\ref{fig:case_eng}. The baseline output has off-target problem while LAVS output generates in the correct language. From the direct token output of LAVS, we can see that many of which are language-specific tokens. Models with LAVS could learn the relation between the target language signal and corresponding language-specific tokens, which further decreases the probability of off-target.

\subsection{Scalability of LAVS} As shown in Table~\ref{tab:ablation}, we explore how the number of language specific(LS) tokens influence the zero-shot performance. The result shows that the OTR keeps decreasing when the number of LS tokens increases. It suggests that more LS tokens can better relieve the off-target issue \textbf{ without harming the supervised performance.}

To test how LAVS generalizes in dataset with more languages, we compare LAVS and VS on OPUS-100~\citep{zhang-etal-2020-improving}. More details of the experiment can be found in Appendix~\ref{app:opus} To alleviate the inference burden, we select all 42 languages with 1M training data for evaluation, which results in 1722 zero-shot directions and 84 supervised directions (en-x and x-en). As shown in Table~\ref{tab:opus}, it turns out that LAVS can  improve the zero-shot performance(-14\% OTR, detailed results in Table~\ref{opus-otr} from appendix) under such setting. Yet, the overall performance is much lower comparing to training on WMT'10. With more languages, the lack of supervision signal would become more problematic for zero-shot translation. LAVS improves the en-x performance by a large margin (+0.9 BLEU, detailed scores in Table~\ref{opus-detail} from appendix), we think separate the vocab of different languages on decoder might have positive influence on general en-x performance.

\begin{table}[t]
    \centering
      \resizebox{0.47\textwidth}{!}{
    \begin{tabular}{ccccc}
\toprule
Shared Tokens(M) & LS Tokens(N) & OTR↓ & Sup. BLEU↑ \\
\midrule 
64k & 0 & 29.4\% &27.5\\
54k & 0 & 33.1\% &26.9\\
54k &10k& 8.2\%  &27.6 \\
54k &20k& 7.4\%  &\textbf{27.8}\\
54k &50k& 5.9\% &27.6\\
54k &212k& \textbf{5\%} &27.6 \\
\bottomrule
\end{tabular}}
    
    \caption{Exploration in number of Language-Specific tokens in LAVS(dec) and the Off-Target Rate on Flores-101. We report the average OTR on zero-shot directions and average BLEU on supervised directions.}
    \label{tab:ablation}
\end{table}

\begin{table}[t]
    \centering
       \resizebox{0.4\textwidth}{!}{
    \begin{tabular}{lcccccc}
        \toprule
        Data & OTR$\downarrow$  & x-y$\uparrow$ & en-x$\uparrow$ & x-en$\uparrow$  \\
        \midrule
        Vocab Sharing & {72\%}  & {1.9}  & {12.6}  & {19.8}    \\
        LAVS (Dec) &\textbf{58\%}  &\textbf{2.3} &\textbf{13.5} &\textbf{20.1} \\
        \bottomrule
    \end{tabular}}
    \caption{Results in OPUS dataset. We evaluate 1722 zero-shot directions and 84 supervised-directions.}
    \label{tab:opus}
\end{table}

\subsection{LAVS's Compatibility with Masked Constrained Decoding} 

\begin{table}[!h]
    \centering
    \resizebox{0.45\textwidth}{!}{
    \begin{tabular}{lccccc}
\toprule
\multirow{2}{*}{Method}  & \multicolumn{2}{c}{DE->CS} & \multicolumn{2}{c}{FR->DE} \\
\cmidrule{2-5}

~&OTR&BLEU &OTR &BLEU \\
\midrule 
Vocab Sharing & 45.1\% & 9.7& 38.3\%&12.7 \\
\quad w/ MCD &30.9\%& 11.4&36.4\%&12.8\\
LAVS (Dec) &18.9\%&13.0&15.4\%&17.2\\ 
\quad w/ MCD &\textbf{11.1\%} &\textbf{14.2}&\textbf{11.3\%}& \textbf{17.8} & \\

\bottomrule
\end{tabular}}
    \caption{The results of masked constrained decoding (MCD) combined with LAVS. Constrained decoding could further improve the performance of LAVS.}
\label{tab:cd}
\end{table}

We propose another method to prevent off-target, which is through masked constrained decoding (MCD). During decoding, the decoder only considers tokens that belong to the target vocab in softmax. The target vocab could be computed using the training corpus. We implement MCD for both original vocab sharing and LAVS. We list the detail of the size of different target vocabs in Table~\ref{tab:vocab-size} from appendix.

As shown in Table~\ref{tab:cd}, it turns out that the method can further improve the zero-shot performance for LAVS (+1.2 BLEU for de-cs, +0.6 BLEU for fr-de). It is worth noticing that, in some direction like FR->DE, the benefit of MCD is rather small for the baseline model (+0.1 BLEU). We think the reason is that the original vocab sharing generates many shared tokens between languages, which will weaken the influence of the constraint. Thus, with more language-specific tokens, LAVS can work better with constrained decoding.

\section{Conclusion}
In this paper, we delve into the hidden reason for the off-target problem in zero-shot multilingual NMT and propose Language-Aware Vocabulary Sharing (LAVS) which could significantly alleviate the off-target problem without extra parameters. Our experiments justify that  LAVS creates a better multilingual vocab than the original Vocabulary Sharing method for multiple languages.


\section{Limitation}
LAVS is proposed to overcome the off-target problem among languages that share alphabets because those languages tend to have more sharing tokens after the sub-word tokenization process. As for language pair that does not have shared tokens, LAVS might not have a direct influence on the zero-shot translation though it can also increase the overall performance for those languages, which might need further exploration.

\section{Acknowledgements}

This paper is supported by the National Key Research and Development Program of China under Grant No.2020AAA0106700 and the National Science Foundation of China under Grant No.61936012. We also thank all reviewers for their valuable suggestions.

\bibliography{anthology,custom}
\bibliographystyle{acl_natbib}

\appendix

\newpage
\section{Method for Completely Separating Vocab}

\begin{figure}[h]
    \centering
    \includegraphics[width=1\linewidth]{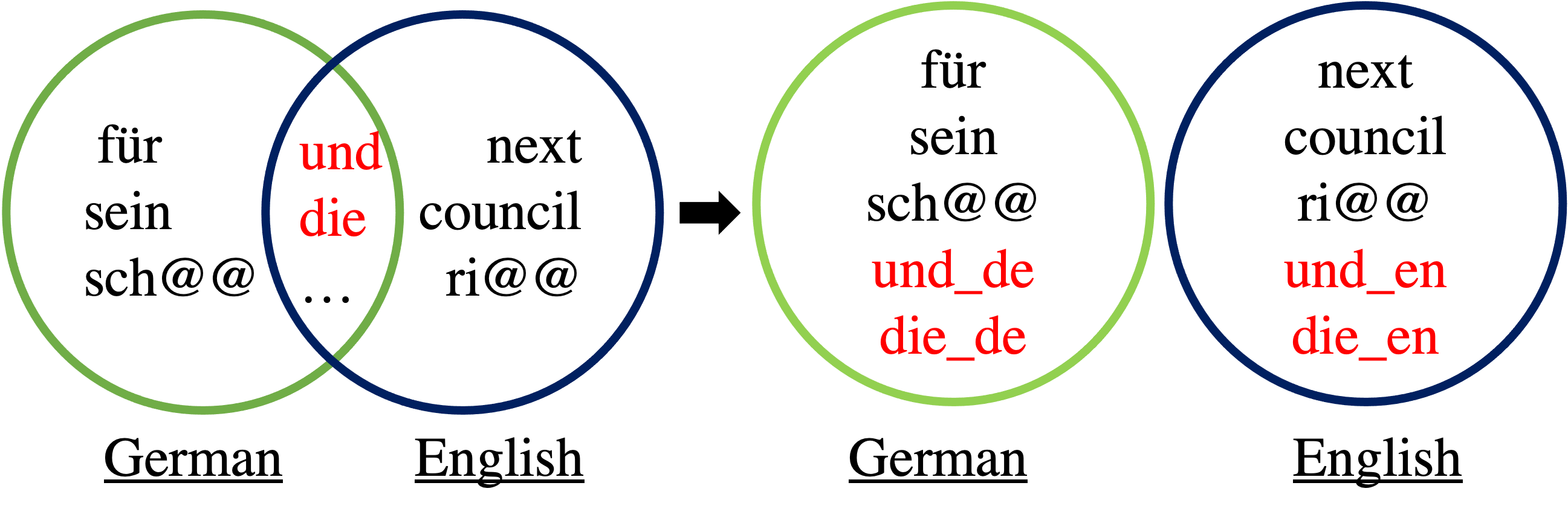}
    \caption{Illustration of completely separating vocabulary of different languages. Note that we don't need to learn a new vocab. Given the original shared vocab, we can split those tokens that are shared by two or more languages into language-specific ones and get a fully separate vocab for each language. }
    \label{fig:complete_split}
\end{figure}

It is easy to turn a shared vocabulary into a separate vocabulary for different languages. As shown in Figure~\ref{fig:complete_split}, we can split the shared token into language specific token if it appears in more than one language.

\section{Separating Tokens by Frequency}
\label{app:alg}

We can also view LAVS from the optimization goal's perspective. We start from only two languages $J$ and $Q$ and compute KL-divergence's change if we only split one shared token to two language-specific tokens. 
\begin{equation}
\begin{aligned}
    \Delta D_{KL}^i & = -J(i)log\frac{J(i)}{Q(i)}-Q(i)log\frac{Q(i)}{J(i)} + \lambda \\
    &= [J(i)-Q(i)]log\frac{Q(i)}{J(i)} + \lambda
\end{aligned}
\label{eq:dkl}
\end{equation}
where we will have two $i$-th tokens for the different languages from the original vocabulary. $\lambda$ is the smoothing factor that can be seen as a constant. According to equation \ref{eq:dkl}, splitting token that has more similar occurrence probability in the two languages will lead to higher increment in language's KL-Divergence (If $J(i)!=Q(i)$, either the $J(i)-Q(i)$ term or the log term will be negative, and the multiply result will also be negative. If $J(i) = Q(i)$ it will be zero, thus reaching the maximum). Also considering the fact that the tokens with high frequency influence the training process much more than the near-zero ones, we should first split the tokens that appear in \textit{two or more} languages all with \textit{high frequency}.

\section{Datasets}
\label{app:dataset}
\subsection{WMT'10}

Following \citet{wang-etal-2020-multi,yang-etal-2021-improving-multilingual,xu-etal-2021-improving-multilingual}, we collect data from freely-accessible WMT contests to form a English-Centric WMT10 dataset.

\begin{table}[h]
    \centering
    \begin{tabular}{cccc}
\toprule
Direction & Train &Test& Dev \\
\midrule 
Fr$\leftrightarrow$En & $10.00 \mathrm{M}$ &newstest15 & newstest13 \\
Cs$\leftrightarrow$En & $10.00 \mathrm{M}$&newstest18 & newstest16 \\
De$\leftrightarrow$En & $4.60 \mathrm{M}$&newstest18 & newstest16 \\
Fi$\leftrightarrow$En & $4.80 \mathrm{M}$&newstest18 & newstest16 \\
Lv$\leftrightarrow$En & $1.40 \mathrm{M}$&newstest17 & newsdev17 \\
Et$\leftrightarrow$En & $0.70 \mathrm{M}$&newstest18 & newsdev18 \\
Ro$\leftrightarrow$En & $0.50 \mathrm{M}$&newstest16 & newsdev16 \\
Hi$\leftrightarrow$En & $0.26 \mathrm{M}$&newstest14 & newsdev14 \\
Tr$\leftrightarrow$En & $0.18 \mathrm{M}$&newstest18 & newstest16 \\
Gu$\leftrightarrow$En & $0.08 \mathrm{M}$&newstest19 & newsdev19 \\
\bottomrule
\end{tabular}
    
    \caption{Description for WMT'10 Dataset.}
    \label{tab:fr-x}
\end{table}

\subsection{Flores-101}

Flores-101~\citep{flores1,flores2} is a Many-to-Many multilingual translation benchmark dataset for 101 languages. It provides parallel corpus for all languages, which makes it suitable to test the zero-shot translation performance of multilingual NMT model. We use the devtest split of the dataset, and only test on the languages that appear during supervised training.

\begin{table}[h]
    \centering
    \begin{tabular}{ccccc}
\toprule
Language & Code & Split &Size \\
\midrule 
French&Fr & devtest &1012 \\
Czech&Cs& devtest &1012 \\
German&De& devtest &1012 \\
Finnish&Fi& devtest &1012 \\
Latvian&Lv& devtest &1012 \\
Estonian&Et& devtest &1012 \\
Romanian&Ro& devtest &1012 \\
Hindi&Hi& devtest &1012 \\
Turkish&Tr& devtest &1012  \\
Gujarati&Gu& devtest &1012 \\
\bottomrule
\end{tabular}
    
    \caption{Description for Flores-101 Dataset.}
    \label{tab:fr-x}
\end{table}

\section{Experiment on OPUS-100 dataset}
\label{app:opus}
OPUS-100\citep{zhang-etal-2020-improving} is an English-centric dataset consisting of parallel data
between English and 100 other languages. We removed 5 languages (An, Dz, Hy, Mn, Yo) from OPUS, since they are not paired with a dev or testset and train the models with all remaining data. The training configuration is the same as our experiment on WMT'10 dataset. The baseline vocab size is 64k. We also implement the baseline model with a larger vocab (256k) but the performance is much lower than the 64k version so we keep the vocab size to 64k. For LAVS, We set the number of language-specific token to 150k instead of 10k because of the increase of languages. We evaluate the supervised and zero-shot performance on Flores-101 dataset. To alleviate the inference burden, we select all 42 languages with 1M training data to conduct zero-shot translation, which forms 1722 zero-shot directions at all. The  ISO code of the evaluated lanugages are "ar, bg, bn, bs, ca, cs, da, de, el, es, et, fa, fi, fr, he, hr, hu, id, is, it, ja, ko, lt, lv,	mk,	ms,	mt,	nl,	no,	pl,	pt,	ro,	ru,	sk,	sl,	sr,	sv,	th,	tr,	uk,	vi,	zh".

\newpage

\newpage

\section{Visualize the language identifiers' representation}

\begin{figure}[h]
    \centering
    Vocabulary Sharing\\
    \includegraphics[width=0.8\linewidth]{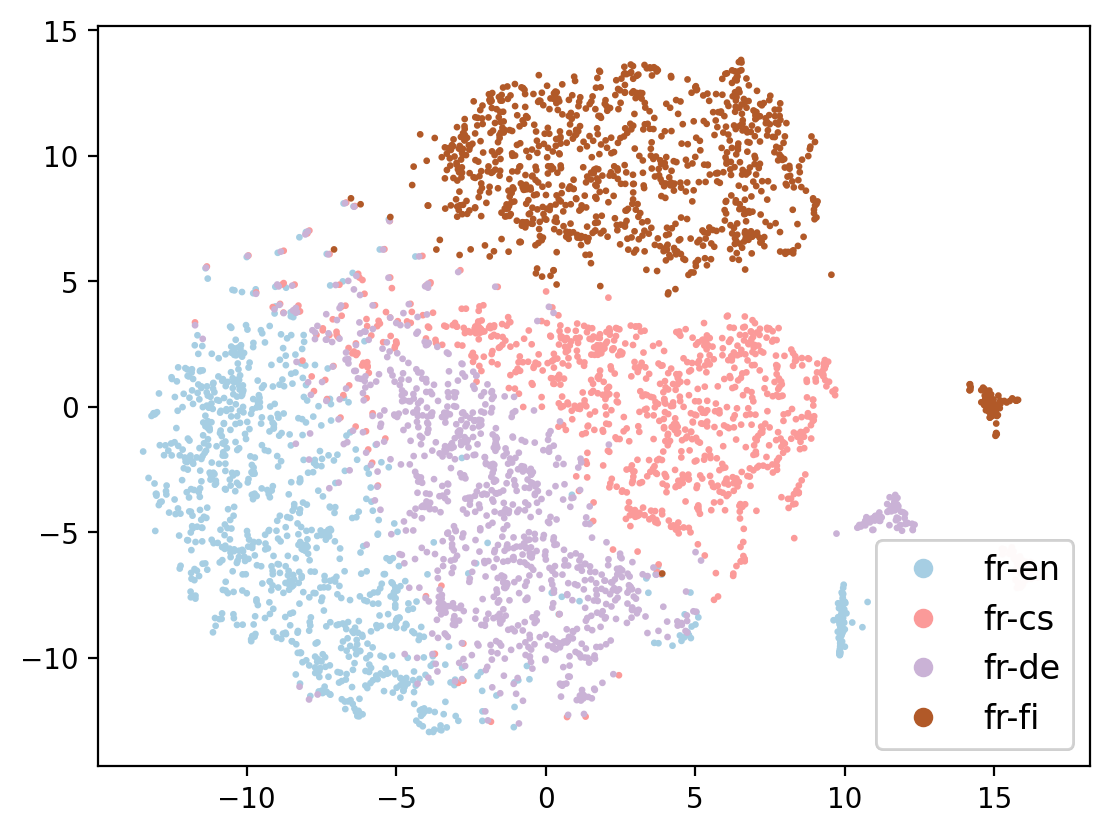}\\
    Language-Aware Vocabulary Sharing\\
    \includegraphics[width=0.8\linewidth]{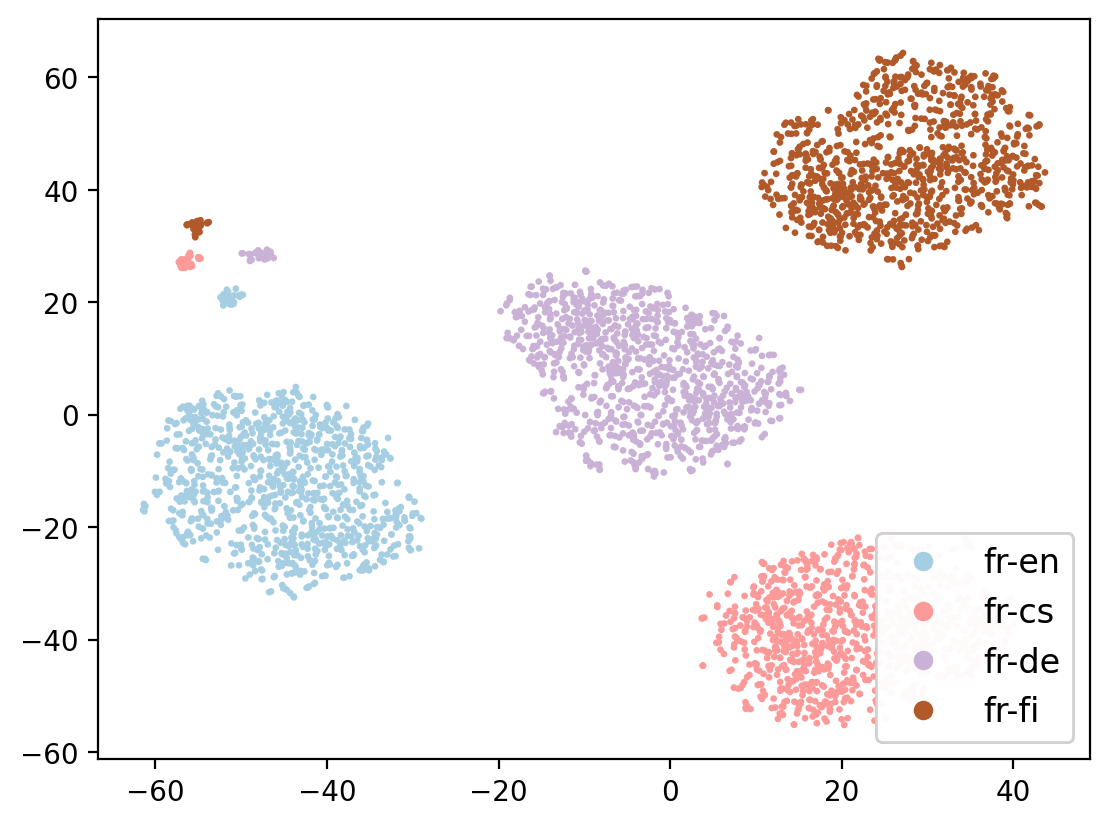}
    \caption{Encoder's hidden output for language identifier token <XX>, visualized using TSNE.}
    \label{fig:token_XX}
\end{figure}

During zero-shot translation, the language identifier token ``<XX>'' is the only element indicating the correct direction. Similar to the visualization in Section~\ref{2.4}, as shown in Figure~\ref{fig:token_XX}, we visualize the <XX> tokens' hidden output(instead of the pooled result from all input tokens) during French-to-Many translation among high-resource languages and compare the results of the original Vocabulary Sharing and LAVS. It turns out that LAVS encodes more discriminative target language information into the <XX> token's hidden output, while the original Vocabulary Sharing fails on that.

In original Vocabulary Sharing the mapping between the target language identifier <XX> and output token is Many-to-One since different language could share output tokens. While for LAVS, the mapping becomes One-to-One for a part of tokens, impulsing the encoder to learn more discriminative representations for the target language identifier.

\newpage

\label{MCD}
\begin{table}[!h]
    \centering
    \begin{tabular}{ccccc}
\toprule
Language & Code &  LAVS & VS \\
\midrule 
French&Fr & 28k & 33k\\
Czech&Cs& 25k & 30k  \\
German&De& 29k & 35k  \\
Finnish&Fi& 23k & 28k\\
Latvian&Lv& 24k & 29k\\
Estonian&Et& 15k & 18k \\
Romanian&Ro& 14k & 20k\\
Hindi&Hi& 10k & 11k \\
Turkish&Tr& 11k &  12k\\
Gujarati&Gu& 7k & 9k\\
\bottomrule
\end{tabular}
    
    \caption{Size of different target vocab for LAVS and VS vocab. Both vocabs have 64k tokens at all. Original VS generally has more tokens in each target vocab, which would weaken the effect of the constrain mask.}
    \label{tab:vocab-size}
\end{table}

\begin{table*}[h]
    \centering
    \resizebox{1.0\textwidth}{!}{
    \begin{tabular}{l|l|l|l|l|l|l|l|l|l|l|l|l|l|l|l|l|l|l|l|l|l|l|l|l|l|l|l|l|l|l|l|l|l|l|l|l|l|l|l|l|l|l|l}
    \hline
            ~ & ar & bg & bn & bs & ca & cs & da & de & el & es & et & fa & fi & fr & he & hr & hu & id & is & it & ja & ko & lt & lv & mk & ms & mt & nl & no & pl & pt & ro & ru & sk & sl & sr & sv & th & tr & uk & vi & zh & AVG \\ \hline

        VS & 0.96 & 0.87 & 0.39 & 0.56 & 0.53 & 0.92 & 0.61 & 0.92 & \textbf{0.91} & 0.80 & 0.73 & 0.61 & 0.86 & 0.76 & 0.69 & 0.58 & 0.89 & 0.42 & 0.85 & 0.83 & 0.58 & \textbf{0.64} & 0.81 & 0.65 & 0.78 & 0.43 & \textbf{0.43} & 0.87 & 0.64 & 0.91 & 0.71 & 0.87 & 0.85 & 0.85 & 0.78 & 0.85 & 0.75 & \textbf{0.63} & 0.48 & 0.83 & 0.47 & \textbf{0.72} & 0.72 \\ \hline

         LAVS & \textbf{0.93} & \textbf{0.67} & \textbf{0.30} & \textbf{0.35} & \textbf{0.51} & \textbf{0.66} & \textbf{0.48} & \textbf{0.76} & 0.96 & \textbf{0.73} & \textbf{0.49} & \textbf{0.47} & \textbf{0.63} & \textbf{0.69} & \textbf{0.68} & \textbf{0.37} & \textbf{0.78} & \textbf{0.27} & \textbf{0.80} & \textbf{0.74} & \textbf{0.51} & 0.68 & \textbf{0.50} & \textbf{0.49} & \textbf{0.57} & \textbf{0.27} & \textbf{0.43} & \textbf{0.64} & \textbf{0.50} & \textbf{0.75} & \textbf{0.63} & \textbf{0.74} & \textbf{0.60} & \textbf{0.54} & \textbf{0.52} & \textbf{0.66} & \textbf{0.57} & 0.74 & \textbf{0.21} & \textbf{0.67} & \textbf{0.47} & 0.77 & \textbf{0.58} \\ \hline
    \end{tabular}}

\caption{Detailed zero-shot OTR of X-to-Many experiment on OPUS-100. Each score denotes the average OTR from X to other 41 languages.}
\label{opus-otr}
    
\end{table*}

\begin{table*}[h]
    \centering
    \resizebox{1.0\textwidth}{!}{
    \begin{tabular}{l|l|l|l|l|l|l|l|l|l|l|l|l|l|l|l|l|l|l|l|l|l|l|l|l|l|l|l|l|l|l|l|l|l|l|l|l|l|l|l|l|l|l|l}
    \hline
        ~&ar & bg & bn & bs & ca & cs & da & de & el & es & et & fa & fi & fr & he & hr & hu & id & is & it & ja & ko & lt & lv & mk & ms & mt & nl & no & pl & pt & ro & ru & sk & sl & sr & sv & th & tr & uk & vi & zh & AVG \\ \hline
        VS & 7.3 & 17.5 & 6.4 & 12 & 24.7 & 12.7 & 25.5 & 16.5 & 9.7 & 17.2 & 10.2 & 4.2 & 7.4 & 27.7 & 6.8 & \textbf{12.2} & 8.6 & 20.2 & 5.3 & 15.8 & 2.0 & 1.6 & 11.1 & 13.7 & 17.2 & 18.5 & 21.1 & 14 & \textbf{19.9} & 7.5 & 26.1 & 15.8 & 12 & 13.6 & 12 & 0.3 & 22.3 & 3 & 6.3 & 5.6 & 13.3 & 6.7 & 12.6 \\ \hline
        LAVS & \textbf{7.7} & \textbf{18.7} & \textbf{6.7} & \textbf{13.6} & \textbf{25.7} & \textbf{12.9} & \textbf{26.2} & \textbf{18.5} & \textbf{11.0} & \textbf{18.0} & \textbf{10.9} & \textbf{4.6} & \textbf{8.3} & \textbf{29.1} & \textbf{7.4} & 12.0 & \textbf{9.4} & \textbf{21.9} & \textbf{6.2} & \textbf{17.4} & \textbf{2.5} & \textbf{2.3} & \textbf{12.0} & \textbf{14.1} & \textbf{17.5} & \textbf{20.5} & \textbf{22.0} & \textbf{14.7} & 17.5 & \textbf{8.1} & \textbf{27.2} & \textbf{16.2} & \textbf{12.5} & \textbf{14.6} & \textbf{12.7 }& \textbf{0.5} & \textbf{23.5} & \textbf{3.7} & \textbf{7.4} & \textbf{8.0} & \textbf{15.3} & \textbf{7.5} & \textbf{13.5} \\ \hline
    \end{tabular}}

    \caption{Detailed BLEU scores of English-to-Many experiment on OPUS-100. }
    \label{opus-detail}
\end{table*}

\end{document}